\documentclass{article}
\usepackage{ijcai26}

\usepackage{times}
\usepackage{soul}
\usepackage{url}
\usepackage[hidelinks]{hyperref}
\usepackage[utf8]{inputenc}
\usepackage[small]{caption}
\usepackage{graphicx}
\usepackage{amsmath}
\usepackage{amsthm}
\usepackage{amssymb}
\usepackage{booktabs}
\usepackage{algorithm}
\usepackage{algorithmic}
\usepackage[switch]{lineno}
\usepackage{multirow}
\usepackage{subcaption}
\usepackage{makecell}
\usepackage{float}

\newtheorem{remark}{Remark}

\usepackage{color}

\newcommand{\colorr}[1]{\textcolor{black}{#1}}

\title{Learning with Foresight: Enhancing Neural Routing Policy via Multi-Node Lookahead Prediction}

\author{
Xia Jiang$^1$
\and
Yaoxin Wu$^{1}$\footnote{Yaoxin Wu is the corresponding author}\and
Yew-Soon Ong$^{2,3}$\and
Yingqian Zhang$^1$\\
\affiliations
$^1$Eindhoven University of Technology \\
$^2$Nanyang Technological University\\
$^3$Agency for Science, Technology and Research (A*STAR)
\emails
x.jiang1@tue.nl,
wyxacc@hotmail.com,
asysong@ntu.edu.sg,
yqzhang@tue.nl
}

\begin{document}

\maketitle

\begin{abstract}
Neural policies have shown promise in solving vehicle routing problems due to their reduced reliance on handcrafted heuristics. However, current training paradigms suffer from a fundamental limitation: they primarily focus on next-node prediction for solution construction, resulting in myopic decision-making that undermines long-horizon planning capacity. To this end, we introduce Multi-node Lookahead Prediction (MnLP), a novel training strategy that extends the supervised learning paradigm to predict multiple future nodes simultaneously. We incorporate causal and discardable MnLP modules that operate exclusively during training, facilitating models to anticipate multi-step decisions while preserving inference-time efficiency. By incorporating multi-depth auxiliary supervision into the loss function, MnLP equips neural policies with the ability of long-range contextual understanding. Experimentally, MnLP outperforms existing training methods, improving the generalization capability of neural policies across various problem sizes, distributions, and real-world benchmarks. Moreover, MnLP can be seamlessly integrated into diverse neural architectures without introducing additional inference overhead.
\end{abstract}

\section{Introduction}

Vehicle routing problems (VRPs), represented by the travelling salesman problem (TSP) and the capacitated vehicle routing problem (CVRP), appear in many real-world industrial scenarios, such as logistics planning~\cite{SAR2023109011}, circuit design~\cite{brophy2014principles}, and robotic systems~\cite{5954127}. Due to their NP-hard nature, solving VRPs (particularly at large scales) requires a trade-off between computational efficiency and solution quality. This challenge is traditionally tackled by heuristics or metaheuristics. However, designing high-quality heuristics typically requires considerable domain expertise and extensive parameter tuning, limiting their practical applicability.

Recently, there has been a growing research interest in developing neural routing policies, which leverage neural networks to learn heuristics directly from data. Specifically, the Transformer architecture~\cite{vaswani2017attention} is commonly used in constructive neural routing policies, which select nodes from the VRP instance sequentially and construct solutions (i.e., tours) in a step-by-step manner~\cite{kool2018attention,hua2025camp_vrp}. They offer high computational efficiency during inference and can match or even surpass the performance of some traditional algorithms~\cite{hottung2025neural}.

Training neural routing policies is typically accomplished using either reinforcement learning (RL)~\cite{kool2018attention,NEURIPS2020_f231f210,hottung2025polynet} or supervised learning (SL)~\cite{luo2023neural,drakulic2023bqnco}. Recent studies have shown that SL approaches exhibit significantly better generalizability on large-scale VRP instances compared to the RL methods~\cite{luo2023neural,drakulic2023bqnco}, highlighting the potential of SL in developing scalable and generalizable neural routing policies. Current SL-based methods, such as the Light-Encoder-and-Heavy-Decoder (LEHD) model, are typically trained using the cross-entropy loss to predict the next node, conditioned on the previously selected node~\cite{luo2023neural}. However, constructing VRP solutions constitutes a long-horizon planning task, often involving hundreds or thousands of nodes. The standard next-node prediction paradigm risks myopic decisions, as the model may fail to capture longer-range context crucial for route construction, thereby leading to a local optimum.

% Inspired by the multi-token prediction framework of DeepSeek-V3~\cite{deepseekai2025}, 
In this paper, we propose a multi-node lookahead prediction (MnLP) strategy for SL routing policies. Specifically, in addition to the standard decoding head that predicts next node, we introduce causal and discardable MnLP modules tasked with predicting future nodes. An auxiliary loss is designed for training these MnLP modules, thereby implementing the lookahead mechanism to enhance the representation learning of the main model. During the training stage, this mechanism encourages the model to account for long-term impact of its decisions by explicit multi-step predictions. At inference time, the MnLP modules are removed, and only the next-node head is used, ensuring no additional computational overhead compared to standard architectures. To demonstrate the effectiveness of MnLP, we conducted extensive experiments and proved its benefits in enhancing neural policies. 

Our main contributions are summarized as follows: 1) We propose a novel architectural design for multi-node lookahead prediction in neural routing policies. The proposed MnLP modules operate exclusively during training, thereby introducing no additional computational overhead during inference; 2) We introduce a multi-depth auxiliary loss that guides the SL process to improve the model's long-horizon planning capability. This enables the learned policy to anticipate future decisions beyond next-node prediction; 3) We conduct extensive experiments demonstrating that our proposed training strategy can effectively enhance neural routing policies in terms of cross-size and cross-distribution generalization, as well as performance on real-world benchmarks.

\section{Related Work}
\label{sec:literature}

\subsection{Neural Routing Policy}
Neural routing policies constitute an important area within neural combinatorial optimization (NCO), and can be categorized into improvement and construction methods.

Improvement methods start from an initial solution and iteratively refine it through a neural network, which learns to optimize sub-problems, applies local search operators, or augments the existing solvers~\cite{9393606,cheng2023select,kim2023learning}. In contrast, construction methods, pioneered by the Pointer Networks~\cite{vinyals2015pointer}, directly generate problem solutions from scratch.  By leveraging Transformer-like architectures and RL, the Attention Model (AM)  proposed by~\cite{kool2018attention} significantly advanced neural routing policies' performance. Building on AM, subsequent work has improved construction-based neural routing policies along several axes: test-time adaptation~\cite{hottung2022efficient,kim2025neural}, enhanced learning strategies~\cite{10387785}, feature refinement~\cite{10160045,11083123}, and multi-task training~\cite{berto2024routefinder,jiang2024bridging,liu2025a}.

Beyond the typical Transformer-like modeling, some efforts have also explored alternative architectures. For example, LEHD~\cite{luo2023neural} introduced a design with a light encoder and a heavy decoder to improve generalizability. Meanwhile, there are also studies that apply large language models (LLMs) to solve VRPs~\cite{jiang2025large,yin2025vitsp}, leveraging the power of large generative models to tackle routing challenges. 

% In parallel, diffusion-based formulations treat routes as sequences to be denoised, using generative modeling to navigate combinatorial spaces~\cite{li2023from,zhao2025disco}, offering a complementary and non-autoregressive pathway.

\subsection{Training Paradigm}

The training paradigms of neural policies broadly fall into two categories: RL and SL. RL-based approaches have achieved strong performance but typically suffer from sparse, delayed rewards and low sample efficiency (particularly on large-scale instances)~\cite{pmlr-v235-kim24o}. To alleviate these limitations, prior work refines the RL pipeline via policy optimization with multiple optima (POMO)~\cite{NEURIPS2020_f231f210}, meta-learning~\cite{omni2023}, and preference-based optimization~\cite{liaobopo}.

% however, these methods largely retain the fundamental credit-assignment issue inherent to long-horizon combinatorial search.

In parallel, SL-based methods have gained traction for their strong cross-scale generalization~\cite{drakulic2023bqnco,luo2023neural}. By constructing partial solutions, they restrict training to a reduced solution space and shift from full-solution construction to high-quality partial routes, thereby complementing RL by trading exploration for data efficiency and transfer. Empirically, SL markedly improves the generalizability of neural routing solvers~\cite{luo2023neural}. However, despite recent refinements~\cite{3694690,luo2025boosting}, most SL methods still use teacher-forced next-node cross-entropy, which optimizes local decisions rather than sequence-level quality and remains prone to exposure bias and error accumulation. We address this issue by extending standard next-node SL to predict multiple future nodes. This process equips the neural policy with stronger foresight during solution construction, thereby improving both performance and generalizability of SL-based routing policies.

\subsection{Multi-token Prediction}

Multi-token prediction (MTP) is a training framework for LLMs in which the model predicts a block of future tokens at each step rather than only the next token, building on prior work in blockwise decoding and non-autoregressive sequence modeling~\cite{stern2018blockwise}. Increasing evidence suggests that pure next-token prediction underfits objectives requiring multi-step planning and long-horizon credit assignment~\cite{bachmannpitfalls}. By supervising multiple future steps jointly, MTP encourages planful intermediate representations, mitigates myopic yet globally suboptimal behaviors induced by teacher forcing, and serves as a regularizer against overfitting to short-range patterns~\cite{qi2020prophetnet}. Empirically, MTP outperforms next-token prediction for LLM pre-training across diverse scales and tasks~\cite{gloeckle2024better,gerontopoulos2025multi}.

DeepSeek-V3 is a prominent real-world deployment of MTP, using it as an auxiliary pre-training objective~\cite{deepseekai2025}. Inspired by this design, we develop MnLP for SL-based routing policies. While conceptually related, we target VRPs rather than language modeling: MnLP omits heavy transformer blocks in MTP and trains over a feasible set that shrinks as nodes are visited instead of a fixed vocabulary, enabling the policy to anticipate multi-step decisions under an evolving decoding process.

\section{Preliminaries}
\label{sec:pre}

\subsection{VRP Formulation}

A VRP instance of size $n$ is defined on a weighted graph $\mathcal{G} = (\mathcal{V}, \mathcal{E})$, where $\mathcal{V} = \{v_i\}_{i=1}^n$ denotes node features (including the depot and customers) and $\mathcal{E} = \{e_{i,j} \mid i, j \in \{1,\ldots,n\},\; i \neq j\}$ is the set of edges. Without loss of generality, the depot is $v_1$. Each edge $e_{i,j}$ has cost $c_{i,j} \ge 0$, forming a cost matrix $C = [c_{i,j}]$. A feasible solution is represented as a node sequence $\tau=(x_1,\ldots,x_k)$, where each $x_\ell\in[1,n]$ indexes node $v_{x_\ell}\in\mathcal{V}$. The total travel cost is: \begin{equation} C(\tau \mid \mathcal{G}) \;=\; \sum_{\ell=1}^{k-1} c_{x_\ell,\, x_{\ell+1}} + c_{x_k,\, x_1}, \end{equation} 
where we add the wrap-around term $c_{x_k,\, x_1}$ because a closed tour is required. Given the feasible set $\Psi$, the objective is: \begin{equation} \tau^*(\mathcal{G}) \;=\; \arg\min_{\tau \in \Psi} \; C(\tau \mid \mathcal{G}) . \end{equation}

The feasible set $\Psi$ encodes all problem-specific constraints. For the TSP, $\Psi$ consists of Hamiltonian cycles that visit each node exactly once. For the CVRP, $\Psi$ comprises a set of routes $\{\tau_r\}_{r=1}^R$, each starting and ending at $v_1$, such that every customer $i\in\mathcal{V}\setminus\{v_1\}$ with demand $d_i$ is served exactly once and the capacity constraint holds on every route: \begin{equation} \sum_{i \in \tau_r \setminus \{1\}} d_i \;\le\; D \quad \text{for all } r\in[1,R]. \end{equation}

We consider Euclidean VRPs, where each node has coordinates $\mathbf{z}_i \in \mathbb{R}^2$ and edge costs are $c_{i,j} = \lVert \mathbf{z}_i - \mathbf{z}_j \rVert_2$.

\subsection{LEHD model with SL}

We select the LEHD model~\cite{luo2023neural} as the backbone and use the proposed MnLP for its training, since: 1) LEHD is a neural policy that is typically trained by SL rather than RL; 2) unlike other SL models such as BQ-NCO~\cite{drakulic2023bqnco}, which learn an unconditional decision process, LEHD generates solutions in an auto-regressive manner, conditioning each decision on previous steps. This causal factorization matches MnLP’s multi-step supervision: the node predicted by the $(k-1)$-th MnLP module enters the decoding context of the $k$-th, yielding a coherent chain of future nodes. Thus, MnLP can supervise several future decisions at once, improving credit assignment across consecutive steps and reducing exposure bias compared with next-step SL paradigm.

LEHD consists of an encoder (e.g., the Shared Encoder in Figure~\ref{fig:framework}), a decoder with $L-1$ attention blocks (e.g., the Main Decoder), and an output head. Given node features $\mathcal{V} = (v_1,\dots,v_n)$ for an instance, a light encoder (a linear layer followed by an attention block) maps them to node embeddings $H^{(E)} = (\mathbf{h}_1^{(0)},\dots,\mathbf{h}_n^{(0)})$. Each attention block comprises a multi-head self-attention layer ($\operatorname{MultiHeadAttn}$) and a feed-forward network ($\operatorname{FFN}$). Further computational details in the attention block are provided in Appendix A.

Instead of training for constructing the complete tour with $n$ nodes, LEHD learns to sequentially construct the solution in $n_p$ steps for a partial solution $(x_1,...,x_{n_p})$, which is sampled from the optimal complete solution. In step $t \in \{1,...,n_p\}$, the decoding process is conditioned on the embeddings of the first selected node $\mathbf{h}_{x_1}^{(0)}$ and the node selected in the previous step $\mathbf{h}_{x_{t-1}}^{(0)}$, which together serve as the decoding context for the current step. More precisely, they interact with $H_a$ through $L-1$ attention blocks:
\begin{equation}
\begin{aligned}
       \widetilde{H}^{(0)} &= \operatorname{concat}(W_1\mathbf{h}^{(0)}_{{x_1}},W_2\mathbf{h}^{(0)}_{{x_{t-1}}},H_a),\\
        \widetilde{H}^{(1)} & = \operatorname{AttentionBlock}(\widetilde{H}^{(0)}),\\
        & \cdots \\
        \widetilde{H}^{(L-1)} & = \operatorname{AttentionBlock}(\widetilde{H}^{(L-2)}),\\
\end{aligned}
\end{equation}
where $H_a =\{\mathbf{h}^{(0)}_i \mid i \in \{1,\ldots,n_p\}\setminus\{x_{1},\ldots,x_{t-1}\} \}$ and $W_1, W_2$ are learnable matrices. The output head of the LEHD model incorporates an attention block, a linear layer with a learnable matrix $W_O$, and a Softmax function to calculate the probabilities $\mathbf{p}_t$ for selecting among all available nodes:
\begin{equation}
\label{eq:main_head}
\begin{aligned}
        \widetilde{H}^{(L)} & = \operatorname{AttentionBlock}(\widetilde{H}^{(L-1)}),\\
        u_i &= \begin{cases}
        W_O\widetilde{\mathbf{h}}_{i}^{(L)}, &\text{$i \notin \{1, 2\}$}\\
           -\infty, &\text{otherwise}
           \end{cases},\\
        \mathbf{p}_t &= \operatorname{Softmax}(\mathbf{u}),
\end{aligned}
\end{equation}

The SL paradigm employs a cross-entropy loss: $\mathcal{L}_{lehd}=-\sum_i^{m}y_i\log(\mathbf{p}_{t,i})$, where $m$ is the number of available nodes; $y_i$ is a binary variable that denotes if $v_i$ is selected. However, if trained only with teacher-forced next-node SL, the model can be locally myopic because auto-regressive decoding weakens credit assignment and accumulates exposure bias, with generalization degrading as $n$ grows.

\section{Methodology}
\label{sec:method}

\begin{figure*}[thb]
\centering
  \includegraphics[width=0.88\textwidth]{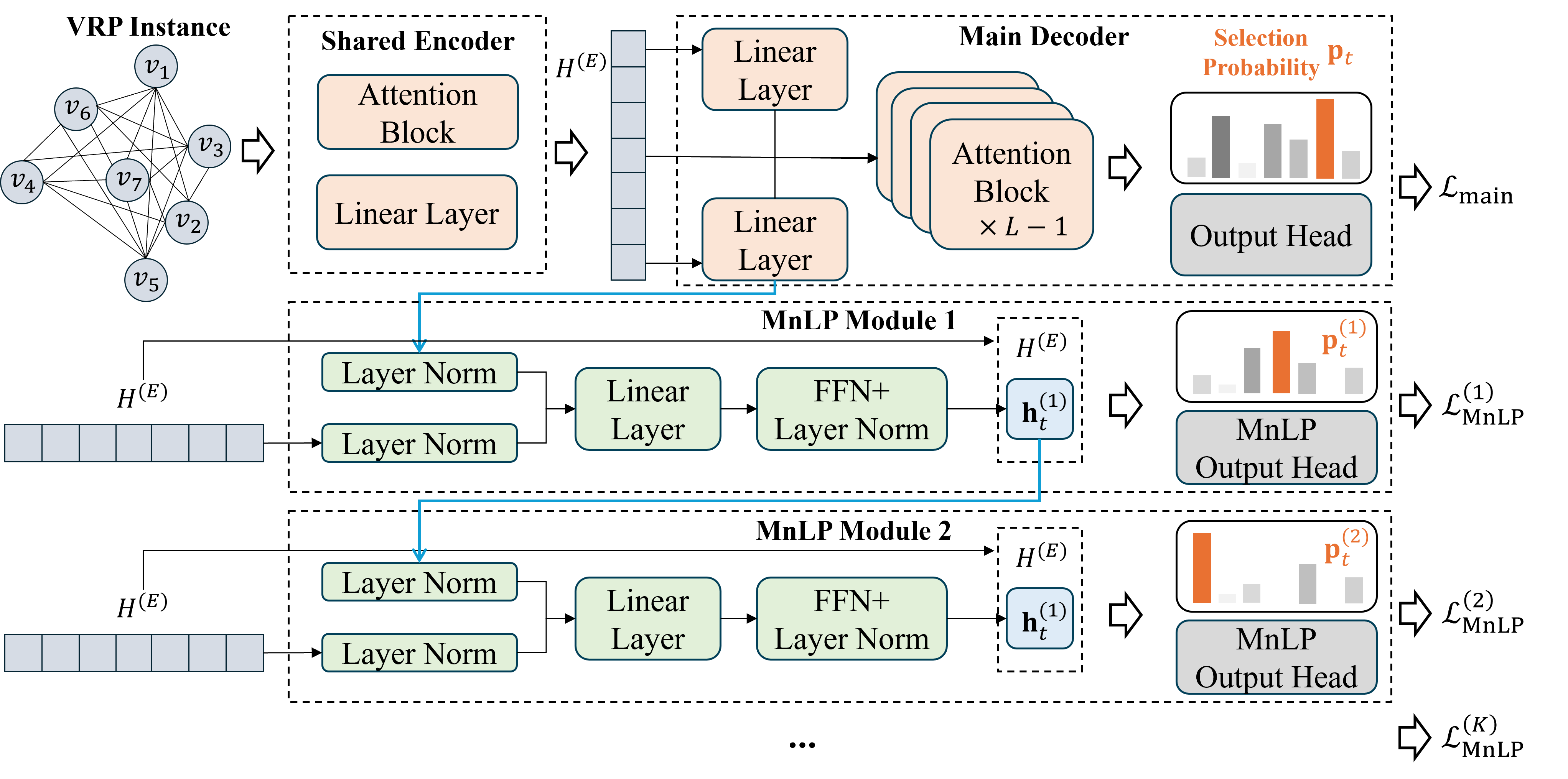}
\caption{\colorr{The overall MnLP model architecture.}}
\label{fig:framework}
\end{figure*}

The MnLP framework (see Figure~\ref{fig:framework}) consists of a shared encoder, a main decoder for next-node prediction, and $K$ causal MnLP modules that predict $K$ future nodes during training.

More specifically, each MnLP module $k \in \{1, \dots, K\}$ predicts node $x_{t + k}$ at decoding step $t$, leveraging the intermediate node representation produced by the preceding module. This design enforces causality across different prediction depths, meaning that each MnLP module receives contextual information from the previous one to predict the corresponding node. From these predictions, we obtain node-selection probabilities $\mathbf{p}^{(k)}_t$ and corresponding losses $\mathcal{L}^{(k)}_{\mathrm{MnLP}}$, which are combined with the main model loss to promote long-horizon planning. Notably, all MnLP modules are used solely during training and are discarded at inference time.

% Notably, all MnLP modules are used solely during training to provide auxiliary supervision. They are discarded at inference time, ensuring that the policy remains lightweight and unchanged for deployment. To this end, they are designed to be lightweight and simple, serving primarily to approximate multi-step causality. In the following subsections, we will describe the model architecture and training details.

\subsection{Model Architecture}

The overall architecture of our model is illustrated in Figure~\ref{fig:framework}. After encoding the VRP instance using the shared encoder, the main decoder and the MnLP modules perform depth-wise computation and apply task-specific output heads to select the node at depth $k$. The process is detailed below.

\textbf{Encoder.} The model uses a lightweight encoder that linearly projects raw node features $\mathcal{V} = (v_1,...,v_n)$ (e.g., 2-D coordinates for Euclidean TSP) and then applies a single Transformer-style attention block (MultiHeadAttn + FFN, as detailed in Appendix A) to produce the node-embedding matrix $H^{(E)}=(\mathbf{h}_1^{(0)},...,\mathbf{h}_n^{(0)})$, such that,
\begin{align}
\tilde{H}^{(E)} &= \mathcal{V}W_E + \mathrm{MultiHeadAttn}(\mathcal{V}W_E),\\
H^{(E)} &= \tilde{H}^{(E)} + \mathrm{FFN}(\tilde{H}^{(E)}),
\end{align}
where $W_E$ denotes the learnable parameter of the linear layer. The node embeddings $H^{(E)}$ are shared by the decoder and all MnLP modules for depth-wise computation.

\textbf{\colorr{Depth-wise Computation.}} The main decoder and all the MnLP modules compute the representations used for node selection. As depicted by Figure~\ref{fig:framework}, the $k$-th MnLP module comprises two LayerNorm ($\operatorname{LN}$) operations, a linear layer with a learnable matrix $W^{(k)}_I$, an $\operatorname{FFN}$ layer, and a task-specific output head. At depth $k=0$, the main model predicts $x_t$ at step $t$ using the embedding of the previously selected node $x_{t-1}$ as part of the decoding context. We generalize this process to a multi-node prediction setting: for any $k>0$, the $k$-th MnLP module predicts node $x_{t+k}$ using an intermediate context representation $\mathbf{h}^{(k)}t$, obtained by combining 1) the representation from the $(k-1)$-th module, $\mathbf{h}^{(k-1)}t$, and 2) the embedding of the ground-truth node $x_{t+k-1}$, $\mathbf{h}^{(0)}_{t+k-1}$ (for $k=1$, $\mathbf{h}^{(k-1)}_t$ reduces to the embedding of $x_{t-1}$). Following the normalization-then-projection design of DeepSeek-V3~\cite{deepseekai2025}, we compute:
\begin{equation}
\label{eq:mlp_rep}
    \mathbf{h}^{'(k)}_t = W^{(k)}_I \operatorname{concat}(\operatorname{LN}(\mathbf{h}^{(k-1)}_t), \operatorname{LN}(\mathbf{h}^{(0)}_{t+k-1}))
\end{equation}

Note that Equation~(\ref{eq:mlp_rep}) is typically for TSP. In the case of other VRPs, we need to incorporate additional constraint information. For example, we introduce the updated vehicle capacity $C^{(k)}$ for CVRP, as the available vehicle capacity must be dynamically adjusted for each prediction depth $k>0$. This is specified by: $\mathbf{h}^{'(k)}_t = W^{(k)}_I \operatorname{concat}(\operatorname{LN}(\mathbf{h}^{(k-1)}_t), \operatorname{LN}(\mathbf{h}^{(0)}_{t+k-1}), W_cC^{(k)})$, where $W_c$ is also a learnable matrix that projects $C^{(k)}$ to the embedding dimension. After that, we apply a $\operatorname{FFN}$ layer with $\operatorname{LN}$ to $\mathbf{h}^{'(k)}_t$ for enriched representation learning, such that,
\begin{equation}
    \mathbf{h}^{(k)}_t = \operatorname{LN}( \operatorname{FFN}(\mathbf{h}^{'(k)}_t))
\end{equation}
As shown in Figure~\ref{fig:framework}, $\operatorname{OutputHead}_k$ takes input as $\mathbf{h}^{(k)}_t$ and generates probability $\mathbf{p}^{(k)}_t$, which is introduced below.

\textbf{\colorr{Task-specific output head.}} The output head in $k$-th MnLP module predicts the $(t+k)$-th node (i.e., $x_{t+k}$). Specifically, we use a task-specific output head $\operatorname{OutputHead}_k$, which is also a parameterized and learnable component, to allow for an interaction between $\mathbf{h}^{(k)}_t$ and the available node embeddings $H^{(E)}$. As a result, the probabilities of selecting each node at prediction depth $k$ are calculated as below,
\begin{equation}
    \mathbf{p}^{(k)}_t = \operatorname{OutputHead}_k(\mathbf{h}^{(k)}_t, H^{(E)})
\end{equation}
Considering the unique problem characteristics, we design task-specific output heads for different VRP variants. For the TSP, we compute the compatibility between the intermediate MnLP representation $\mathbf{h}^{(k)}_t$ at prediction depth $k$ and the node embeddings $H^{(E)}$ simply via a dot-product operation:
\begin{equation}
    \text{logits} = W^{(k)}_{M}\mathbf{h}^{(k)}_t \cdot (W^{(k)}_{E}{H^{(E)}})^\top, 
\end{equation}
where $W^{(k)}_{M}$ and $W^{(k)}_{E}$ are learnable matrices. This is followed by masking (as remarked below) and a Softmax operation imposed on the feasible set $\mathcal{A}^{(k)}_t$ at each depth $k$.
\colorr{\begin{remark}[Feasible set $\mathcal{A}^{(k)}_t$ at depth $k$]
Let $S_p$ be the node set of a sampled partial tour of length $n_p$. 
At each decoding step $t$, the available nodes (i.e., feasible set) for the depth-$k$ prediction are $\mathcal{A}^{(k)}_t \;=\; \bigl\{\, i \in S_p \setminus \{x_1,\dots,x_{t+k-1}\}\,\bigr\}$.
\end{remark}}

We ensure feasibility by assigning $-\infty$ to the logits of all nodes not in $\mathcal{A}^{(k)}_t$ (same masking as in prior work~\cite{kool2018attention}), including nodes visited by earlier MnLP depths. Consequently, the size of the feasible set
$|\mathcal{A}^{(k)}_t| = n_p - (t+k-1)$ decreases with the increase of $k$. If $t+k>n_p$, we skip the computation of depth-$k$ loss.
Therefore, the final selection distribution is then computed only over $\mathcal{A}^{(k)}_t$:
\begin{equation}
\mathbf{p}^{(k)}_t(i) =
\begin{cases}
\dfrac{\exp(\text{logits}(i))}{\sum\limits_{j \in \mathcal{A}^{(k)}_t} \exp(\text{logits}(j))}, & \text{if }i \in \mathcal{A}^{(k)}_t,\\[1.2ex]
0, & \text{otherwise.}
\end{cases}
\end{equation}

In contrast, the output head of CVRP requires a more elaborate mechanism due to the need to predict both the next customer and the possibility of returning to the depot~\cite{luo2023neural}. 
To address this, we adopt an architecture similar to the decoder layer of the LEHD. Specifically, 
we first concatenate the representation $\mathbf{h}^{(k)}_t$ with $H^{(E)}$ (the embeddings of available nodes): $\widetilde{H}^{(k)} = \operatorname{concat}(\mathbf{h}^{(k)}_t, H^{(E)})$, which is then processed by an attention block to make the decoding context (i.e., $\mathbf{h}^{(k)}_t$) interact with the node embeddings. After that, the output $\widetilde{H}^{(k)}$ is used to compute the node selection probabilities at depth $k$, as formulated similarly by Equation~(\ref{eq:main_head}).

% To enforce tour validity, we apply masking to prevent revisiting cities already included in the current partial tour, which is similar to the main model of the LEHD. This is achieved by constructing a binary mask over the selected cities, setting their corresponding logits to $-\inf$.

\begin{algorithm}[tb]
\caption{Training with MnLP}
\label{alg:mtp_training}
\textbf{Input}: episodes per epoch $E$, batch size $B$ \\
\textbf{Param}: MnLP weight $\gamma$, warm-up epochs $W$, ratio $\alpha$ \\
\textbf{Output}: trained model $\theta$
\begin{algorithmic}[1]
\FOR{epoch $e = 1,2,\dots$}
    \STATE $\gamma_e \leftarrow \gamma \cdot \min\!\bigl(1, e / (\alpha W)\bigr)$
    \FOR{$i = 1$ to $E/B$}
        \STATE Sample a batch of $B$ instances and reset env
        \FOR{decoding step $t = 0,1,\dots$ until termination}
            \IF{$t \le 1$}
                \STATE Select fixed nodes (e.g., depot/start)
            \ELSE
                \STATE $(\mathcal{L}_{\text{main}}, \mathcal{L}_{\text{MnLP}}) \leftarrow \textsc{Forward}(\theta)$
                \STATE $\mathcal{L} \leftarrow \mathcal{L}_{\text{main}} + \gamma \mathcal{L}_{\text{MnLP}}$
                \STATE Update $\theta$ by backprop on $\mathcal{L}$
            \ENDIF
            \STATE $\textsc{StepEnv}()$
        \ENDFOR
    \ENDFOR
\ENDFOR
\STATE \textbf{return} $\theta$
\end{algorithmic}
\end{algorithm}

\subsection{MnLP Training}

MnLP enhances policy learning with auxiliary supervision. As shown in Algorithm~\ref{alg:mtp_training}, each forward pass produces both the main logits for next-node prediction and lookahead logits for MnLP. The main loss $\mathcal{L}_{\mathrm{main}}$ and auxiliary loss $\mathcal{L}_{\mathrm{MnLP}}$ are computed in parallel (Figure~\ref{fig:framework}), and parameters are updated using their weighted sum. Only the main decoder is used at inference, so test-time complexity remains unchanged.

More precisely, we extend the standard cross-entropy used in SL training to supervise predictions at multiple depths. Let $m^k = |\mathcal{A}^{(k)}_t|$ be the size of the feasible set, $\mathbf{p}^{(k)}_{t}\in \mathbb{R}^{m^k}$ be the predicted distribution over $\mathcal{A}^{(k)}_t$, and $y_i^k\in\{0,1\}$ be a one-hot indicator of the node selected at depth $k$ (such that $\sum_i y_i^k = 1$). For a given $k$, the depth-$k$ MnLP loss is:
\begin{equation}
 \mathcal{L}^{(k)}_{\mathrm{MnLP}} =
 \begin{cases}
 -\displaystyle\sum_{i=1}^{m^k} y_i^k \log\!\big(\mathbf{p}^{(k)}_{t,i}\big), & \text{if } t{+}k \le n_p,\\[6pt]
 0, & \text{otherwise},
 \end{cases}
\end{equation}
where $n_p$ is the length of the teacher-forced partial solution. Note that when the lookahead would exceed the partial trajectory, the loss is masked out. To obtain the overall signal, we average the depth-wise losses across the $K$ MnLP modules:
\begin{equation}
    \mathcal{L}_{\mathrm{MnLP}} =
    \frac{1}{K} \sum_{k=1}^K \mathcal{L}^{(k)}_{\mathrm{MnLP}}
\end{equation}
This auxiliary loss is then integrated into the training objective via a weighted sum with the main task loss:
\begin{equation}
    \mathcal{L} = \mathcal{L}_{\mathrm{main}} + \gamma \mathcal{L}_{\mathrm{MnLP}},
\end{equation}
Here, $\gamma$ is the parameter that controls the contribution of the MnLP auxiliary supervision, and $\mathcal{L}_{\mathrm{main}}$ denotes the cross-entropy loss of the main model. To prevent MnLP from dominating early training, we linearly ramp up $\gamma$ over a warm-up period of $W$ epochs (line 2 in Algorithm~\ref{alg:mtp_training}).

We train with $\mathcal{L}$ that augments the standard next-node head with $K$ lookahead heads ($k{=}1,\dots,K$). At depth $k$, supervision targets the node that would appear at step $t{+}k$ under teacher forcing, but is applied already at $t$: the $k$-th MnLP module is conditioned on $x_{t+k-1}$ and its cross-entropy is computed over $\mathcal{A}^{(k)}_t$. This design shortens credit-assignment paths, strengthens long-range representations in the encoder, and provides multi-task regularization during training.

\renewcommand{\arraystretch}{0.8}

\begin{table*}[t!]
    \centering
    \setlength{\tabcolsep}{1mm}
    {\fontsize{9}{11}\selectfont
    \begin{tabular}{llccccccccccccc}
        \toprule
        \multirow{13}{*}[-5ex]{\rotatebox{90}{\emph{TSP}}} 
        &  & \multicolumn{3}{c}{$n=100$} & \multicolumn{3}{c}{$n=200$} 
        & \multicolumn{3}{c}{$n=500$} & \multicolumn{3}{c}{$n=1000$} \\
        & & Obj. & Gap & Time & Obj. & Gap & Time 
          & Obj. & Gap & Time & Obj. & Gap & Time \\
        \midrule
        & Concorde   & 7.763 & - & 34m  & 10.704 & - & 3m 
                    & 16.522 & - & 32m & 23.120 & - & 7.8m \\
        & OR-Tools  & 7.947 & 2.368\% & 11h & 11.091 & 3.618\% & 17m 
                    & 17.296 & 4.682\% & 50m & 24.249 & 4.885\% & 10h \\ 
        \cmidrule(lr){2-14}
        & POMO augx8    & 7.773  & 0.134\% & 0.5m 
                        & 10.868  & 1.533\% & 2.5s 
                        & 20.188  & 22.187\% & 0.4m 
                        & 32.500 & 40.570\% & 0.7m  \\
        & BOPO augx8  & 7.771 & 0.103\% & 0.5m  
                      & 10.880 & 1.644\% & 2.5s 
                      & 19.109 & 15.658\% & 0.4m 
                      & 29.571 & 27.902\% & 0.7m \\
        & SA-DABL augx8  
                      & 7.767 & 0.053\% & 0.5m  
                      & 10.853 & 1.392\% & 2.5s 
                      & 19.034 & 15.204\% & 0.4m 
                      & 29.681 & 28.378\% & 0.7m \\
        & ELG augx8  & 7.781 & 0.232\% & 1.3m  
                     & 10.854 & 1.401\% & 3.1s 
                     & 17.714 & 7.215\% & 0.2m 
                     & 25.763 & 11.432\% & 0.7m \\
        & INViT greedy  & 7.907 & 1.855\% & 1.6m  
                         & 11.079 & 3.503\% & 4.7s 
                         & 17.392 & 5.266\% & 13s 
                         & 24.578 & 6.306\% & 0.7m \\
        & DGL greedy  & 7.805 & 0.531\% & 1.9m  
                       & 10.839 & 1.261\% & 7.7s 
                       & 16.899 & 2.282\% & 19.6s 
                       & 23.739 & 2.678\% & 0.6m \\
        & LEHD greedy   & 7.807 & 0.558\% & 9.6s  
                         & 10.792 & 0.824\% & 1.2s 
                         & 16.810 & 1.748\% & 6.6s 
                         & 23.944 & 3.563\% & 1.3m \\
        \cmidrule(lr){2-14}
        & Ours (MnLP) greedy  
                & 7.805 & 0.531\% & 9.6s  
                & 10.789 & 0.795\% & 1.2s 
                & 16.804 & 1.711\% & 6.6s 
                & 23.779 & 2.852\% & 1.3m \\
        & \quad RRC 100  
                & 7.764 & 0.011\% & 4.6m  
                & 10.710 & 0.059\% & 0.7m 
                & 16.576 & 0.333\% & 4.0m 
                & 23.375 & 1.105\% & 19m \\
        & \quad RRC 500  
                & 7.763 & 0.002\% & 22m  
                & 10.707 & 0.026\% & 3.4m 
                & 16.556 & 0.207\% & 16m 
                & 23.290 & 0.736\% & 1.5h \\
        & \quad RRC 1000  
                & 7.763 & 0.001\% & 42m  
                & 10.705 & 0.017\% & 6.5m 
                & 16.550 & 0.174\% & 31m 
                & 23.268 & 0.639\% & 2.7h \\
        \midrule

        % ---------------- CVRP ----------------
        \multirow{14}{*}[-1ex]{\rotatebox{90}{\emph{CVRP}}} 
        & LKH3  & 15.647 & - & 12h 
                 & 20.173 & - & 2.1h 
                 & 37.229 & - & 5.5h 
                 & 37.091 & - & 7.1h \\
        & HGS  & 15.564 & -0.533\% & 4.5h 
                & 19.946 & -1.126\% & 1.4h 
                & 36.561 & -1.794\% & 4.0h 
                & 36.289 & -2.162\% & 5.3h \\
        & OR-Tools  & 16.616  & 6.193\% & 2h 
                     & 21.564 & 6.894\% & 1h 
                     & 40.698 & 9.112\% & 2.2h 
                     & 41.417  & 11.662\% & 3h \\ 
        
        \cmidrule(lr){2-14}
        & POMO augx8    
                & 15.755  & 0.689\% & 0.5m 
                & 21.155  & 4.866\% & 2.9s 
                & 44.638  & 19.901\% & 0.5m 
                & 84.896 & 128.885\% & 0.9m \\
        & SA-DABL augx8    
                & 15.906  & 1.655\% & 0.5m 
                & 20.980  & 3.999\% & 2.9s 
                & 41.869  & 12.463\% & 0.5m 
                & 51.478 & 38.788\% & 0.9m \\
        & ELG augx8   
                & 15.839 & 1.227\% & 1.0m  
                & 20.699 & 2.608\% & 5.0s 
                & 39.388 & 5.799\% & 0.3m 
                & 41.548 & 12.016\% & 1.0m \\
        & RELD augx8   
                & 15.797 & 0.960\% & 0.7m  
                & 20.506 & 1.654\% & 2.4s 
                & 38.337 & 2.975\% & 7.8s 
                & 39.597 & 6.757\% & 0.5m \\
        & INViT greedy  
                & 17.206 & 9.964\% & 3.3m  
                & 22.626 & 12.160\% & 8.1s 
                & 42.356 & 13.772\% & 19s 
                & 42.858 & 15.548\% & 1.3m \\
        & DGL greedy  
                & 16.654 & 6.441\% & 14m  
                & 21.861 & 8.368\% & 0.5m 
                & 40.712 & 9.356\% & 1.4m 
                & 43.295 & 16.727\% & 2.9m \\
        & LEHD greedy   
                & 16.233 & 3.748\% & 0.4m  
                & 20.809 & 3.156\% & 2.4s 
                & 38.359 & 3.035\% & 7.2s 
                & 39.877 & 7.513\% & 1.3m \\
        \cmidrule(lr){2-14}
        & Ours (MnLP) greedy  
                & 16.196 & 3.509\% & 0.4m  
                & 20.819 & 3.206\% & 2.4s 
                & 38.319 & 2.928\% & 7.2s 
                & 39.388 & 6.195\% & 1.3m \\
        & \quad RRC 100  
                & 15.720 & 0.471\% & 5.7m  
                & 20.245 & 0.363\% & 3.2m 
                & 37.462 & 0.627\% & 6.6s 
                & 38.151 & 2.859\% & 16m \\
        & \quad RRC 500  
                & 15.661 & 0.094\% & 26m  
                & 20.146 & -0.136\% & 4.0m 
                & 37.231 & 0.007\% & 17m 
                & 37.693 & 1.624\% & 1.3h \\
        & \quad RRC 1000  
                & 15.648 & 0.012\% & 51m  
                & 20.121 & -0.255\% & 7.9m 
                & 37.133 & -0.257\% & 34m 
                & 37.516 & 1.148\% & 2.7h \\
        \bottomrule
    \end{tabular}}
    \caption{Experimental results on TSP and CVRP instances, with optimality gaps computed using Concorde for TSP and LKH3 for CVRP.}
    \label{tab:gaps}
\end{table*}

\section{Experiments}
\label{sec:exp}

We empirically evaluate our proposed MnLP training strategy on TSP and CVRP
of various sizes and distributions, comparing it against the existing neural routing policies with different architectures and training algorithm designs.

\paragraph{Problem setting}We follow standard data-generation protocols for TSP and CVRP in prior work~\cite{kool2018attention}. In line with the settings used by LEHD~\cite{luo2023neural}, we use one million instances each for TSP100 (i.e., TSP instance with 100 nodes) and CVRP100 as the training sets. For evaluation, the TSP test set comprises 10{,}000 instances with 100 nodes, along with 128 instances each for problem sizes of 200, 500, and 1000 nodes. The CVRP test set mirrors this structure, containing the same number of instances across corresponding problem sizes. To obtain ground-truth labels, the optimal solutions for the TSP training set are computed using the Concorde solver~\cite{cook2011traveling}, while optimal solutions for the CVRP training set are generated by the Hybrid Genetic Search (HGS) solver~\cite{vidal2022hybrid}.

\paragraph{Model and training setting}We adopt the LEHD configuration from~\cite{luo2023neural}, which has a one-layer encoder and a decoder with 6 attention blocks. The node embedding dimension is set to 128. Each $\operatorname{MultiHeadAttn}$ layer uses 8 attention heads, while the dimension of $\operatorname{FFN}$ is 512. We set $K=4$ for MnLP, while a sensitivity analysis of $K$ is also conducted. The MnLP weight $\gamma$ is 0.2 for TSP and 0.1 for CVRP, with $W=5$ warm-up epochs and warm-up ratio $\alpha=3$. TSP and CVRP models are trained for 150 and 15 epochs, respectively, with a batch size of 1024. We use Adam with an initial learning rate $10^{-4}$ and decay rates 0.97 for TSP and 0.9 for CVRP. All experiments are run on a server with an AMD EPYC 7F72 CPU and an NVIDIA H100 GPU.

\paragraph{Baselines}We compare with: \textbf{(1) Classical
solvers: }Concorde~\cite{cook2011traveling}, LKH3~\cite{helsgaun2017extension}, HGS~\cite{vidal2022hybrid}, and OR-Tools~\cite{ortools_routing}; \textbf{(2) Neural routing policies with architectural designs: } ELG~\cite{gao24ijcai}, LEHD~\cite{luo2023neural}, INViT~\cite{fang24c}, DGL~\cite{xiao2025dgl}, and RELD~\cite{huang2025rethinking}; \textbf{(3) Neural routing policies with training algorithm designs: }POMO~\cite{NEURIPS2020_f231f210}, which train using parallel solution trajectories, SA-DABL~\cite{3694690}, which implements SL with data augmentation and bidirectional loss, and BOPO~\cite{liaobopo}, which uses best-anchored and objective-guided preference optimization. We retrain LEHD by following ~\cite{luo2023neural}, and use the provided open-sourced models for other baselines. During evaluation, we measure: 1) average objective values (Obj.), 2) performance gap relative to baseline solvers (Gap), and 3) total inference time (Time).

\subsection{Main Results}

Table~\ref{tab:gaps} reports results on uniformly distributed routing instances of various sizes. MnLP policies consistently outperform the original greedy LEHD, with gains increasing at larger scales. For example, on TSP1000, MnLP reduces the optimality gap from 3.563\% to 2.852\%—a ~20\% relative improvement, showing that multi-step supervision enhances long-horizon decision quality. These gains arise purely from training (MnLP modules are not used at inference), isolating the benefit to representation learning. Overall, the results support our claim that supervising multiple future steps yields richer long-range context and smaller gaps, especially on large instances where myopic errors accumulate.

More broadly, MnLP can also outperform other training strategies in the domain of NCO, including POMO, preference-based optimization (i.e., BOPO), and augmentation-driven SL (i.e., SA-DABL), especially when the instance size exceeds $100$. The performance gain is more pronounced on $n = 1000$ instances, suggesting that MnLP training is particularly effective for large-scale problems. Meanwhile, compared to methods that introduce additional components and computationally intensive processes (e.g., INViT and DGL), MnLP achieves better efficiency while producing more optimal solutions across most task settings.

Although some models (e.g., ELG) perform better on in-distribution $n=100$ cases, MnLP remains competitive and can be strengthened via the Random Re-Construct (RRC) process from LEHD~\cite{luo2023neural}. With only 100 RRC steps, MnLP surpasses these compared policies. Meanwhile, we also present the comparative results between our method and LEHD with the same RRC steps in Appendix C, which shows that our model with RRC can also outperform the original model, particularly on large-scale instances, demonstrating compatibility with test-time refinement techniques.

\renewcommand{\arraystretch}{0.8}

\begin{table}[!t]
  \centering
  \begin{tabular}{ll|cc|cc}
    \toprule
    \multirow{2}{*}{ } & \multirow{2}{*}{Method} & \multicolumn{2}{c|}{$n=500$} & \multicolumn{2}{c}{$n=1000$} \\
    \cmidrule(lr){3-6}
    & & $R$ & $E$ & $R$ & $E$ \\
    \midrule
    \multirow{2}{*}{TSP} & LEHD & 2.888 & 2.821 & 5.937 & 6.397 \\
    & MnLP & 2.304 & 2.447 & 4.458 & 5.189 \\
    \midrule
    \multirow{2}{*}{CVRP} & LEHD & 4.383 & 4.558 & 8.188 & 8.972 \\
    & MnLP & 4.294 & 4.518 & 7.831 & 8.843 \\
    \bottomrule
  \end{tabular}
  \caption{Generalization performance comparison: Average gaps (\%) across 1,000 instances for rotation and explosion distributions.}
  \label{tab:generalization}
\end{table}

\begin{table}[thbp]
\centering
\setlength{\tabcolsep}{1.6mm}
\renewcommand{\arraystretch}{0.8}
{\fontsize{9}{11}\selectfont
\begin{tabular}{c| l |c|c|c}
\toprule
&        & POMO & LEHD & Ours (MnLP)  \\ 
& Size  & aug$\times$8 & greedy & greedy\\
\midrule
\multirow{6}{*}{\rotatebox{90}{TSPLib}}& $<$100   & 0.792\% & 1.064\%  & 0.673\%   \\
&100-200 & 2.423\% & 2.309\%  & 2.127\%  \\
&200-500  & 13.413\% & 4.371\%  & 2.414\%  \\
&500-1k   & 31.678\% & 8.729\%  & 15.597\%  \\
& $>$1K  & 63.810\% & 13.913\% & 12.252\%  \\
\midrule
&All  & 26.439\% & 6.841\% & 6.401\%  \\
\midrule
% &         & POMO & LEHD & \multicolumn{2}{c}{LEHD (MnLP)}  \\ 
% &Set (size)  & aug$\times$8 & greedy & greedy & RRC\\
&Set (size)  &  &  &  \\
\hline
\multirow{7}{*}{\rotatebox{90}{CVRPLib}}
&A  (31-79)     & 4.970\% & 6.595\%  & 7.114\%   \\
&B  (30-77)     & 4.747\% & 7.493\%  & 7.137\%  \\
&E  (12-100)    & 11.402\% & 5.478\%  & 4.791\%   \\
&F  (44-134)    & 15.973\% & 8.491\%  & 7.391\%  \\
&M  (100-199)   & 4.861\% & 8.407\%  & 8.197\%   \\
&P  (15-100)    & 15.525\% & 6.061\%  & 6.564\%  \\
&X  (100-1k)    & 21.684\% & 11.178\% & 9.024\%   \\
\hline
& All  & 15.450\% & 9.038\% & 7.945\% \\
\bottomrule
\end{tabular}
}
\caption{Experimental results on TSPLib and CVRPLib.}
\label{tab:lib}
\end{table}

\subsection{Cross-distribution Generalization}

Generalizability is essential for neural routing policies: models that overfit a single distribution have limited practical value~\cite{omni2023}. Beyond the performance on uniformly distributed instances reported in Table~\ref{tab:gaps}, we further evaluate the cross-distribution generalization performance of our model on two typical benchmark distributions of VRP: the \textit{rotation distribution} ($R$) and the \textit{explosion distribution} ($E$), both of which have been used extensively in prior work~\cite{omni2023,NEURIPS2023_a68120d2}. The precise generation protocols are given in Appendix B.

Specifically, we randomly generate 1,000 instances for each distribution with different sizes: $n = 500$ and $1000$. The comparative results are summarized in Table~\ref{tab:generalization}. It shows that the model trained with MnLP consistently improves upon the LEHD baseline across distributions and problem sizes. \colorr{Taking TSP as an example, MnLP yields substantial relative reductions in average gap: at $n=1000$, $R$ decreases from $5.937\%$ to $4.458\%$ ($24.9\%$ improvement) and $E$ from $6.397\%$ to $5.189\%$ ($18.9\%$ improvement).} These results indicate that MnLP can effectively enhance cross-distribution generalizability and robustness to distributional shift.

Furthermore, we evaluate the MnLP-trained model on the widely used real-world benchmark datasets, including TSPLib~\cite{tsplib} and CVRPLib~\cite{uchoa2017new}, which reflect real-world problem distributions. The performance is evaluated on instances with Euclidean distance, and is compared against POMO and the original LEHD model in Table~\ref{tab:lib}. Empirically, we observe that the MnLP improves overall performance across the benchmark datasets, showing its effectiveness on instances with various distributions. 

\begin{figure}[t!]
\centering
    \subfloat[]{\includegraphics[width=0.24\textwidth]{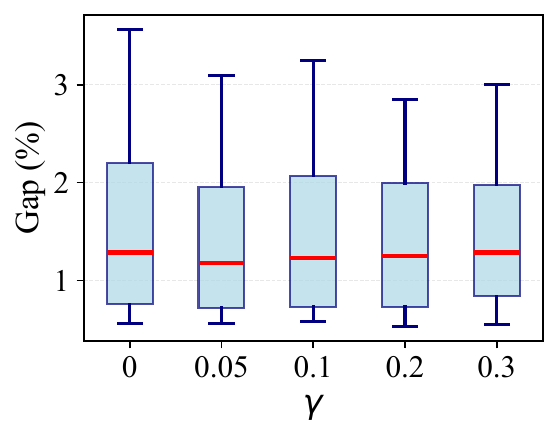}}\hfill
    \subfloat[]{\includegraphics[width=0.24\textwidth]{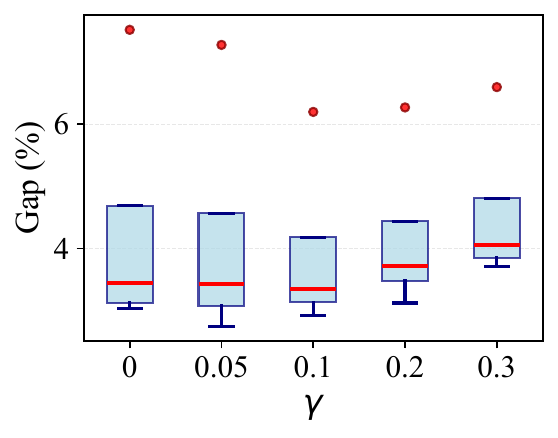}}
\caption{The distribution of optimality gap across different problem sizes for varying values of $\gamma$. (a) TSP; (b) CVRP.}
\label{fig:gamma}
\end{figure}

\begin{table}[thbp]
\centering
\setlength{\tabcolsep}{1.6mm}
{\fontsize{9}{11}\selectfont
\begin{tabular}{c|l|c|c|c|c}
\toprule
 \multicolumn{2}{c|}{Size ($n$)}       & $100$ & $200$ & $500$ & $1000$  \\ 
\midrule
\multirow{2}{*}{TSP}
&MnLP-E   & 0.531\% & 0.795\% & 1.711\%  & 2.852\%  \\
&MnLP-D    & 0.572\%& 0.804\%  & 1.823\%  & 3.442\%  \\
\midrule
\multirow{2}{*}{CVRP}
&MnLP-E   & 3.509\%& 3.206\%  & 2.928\%  & 6.195\%  \\
&MnLP-D     & 4.008\%& 3.332\%  & 3.215\%  & 7.286\%  \\
\bottomrule
\end{tabular}
}
\caption{The ablation study of the MnLP implementations.}
\label{tab:mnlp_structure}
\end{table}

\subsection{Ablation and Sensitivity Analyses}

We conduct ablation and sensitivity analyses to identify the optimal MnLP training configuration.

\paragraph{Effect of the value of $\gamma$.}The impact of different values of $\gamma$ on model performance is shown in Figure~\ref{fig:gamma}. By varying $\gamma$ from $\{0.05,0.1,0.2,0.3\}$, we observe that appropriate values of $\gamma$ lead to improved training outcomes. For instance, setting $\gamma = 0.2$ for TSP and $\gamma = 0.1$ for CVRP yields the best average performance across various problem sizes. In contrast, excessively large values (e.g., $\gamma=0.3$) may degrade performance, as the training process becomes increasingly biased toward optimizing the MnLP modules, thereby hindering sufficient learning of the main model for next node prediction.

\paragraph{Effect of different MnLP implementations.}To evaluate our design, we compare different architectural implementations. In our method, MnLP modules share the encoder with the main model. Alternatively, they can share high-level node representations by reusing the network components before the output head, where the first MnLP module takes input from the $(L-1)$-th decoder layer. We denote our encoder-sharing design as MnLP-E and the decoder-based variant as MnLP-D, and report results in Table~\ref{tab:mnlp_structure}. MnLP-E consistently outperforms MnLP-D, while MnLP-D sometimes even degrades performance compared with the original LEHD. This is likely because decoder representations already contain the current decoding context (e.g., the node selected at the previous step), which introduces noise for future prediction.

\paragraph{Effect of other factors.} We further analyze the computational cost of MnLP, the impact of using different numbers of MnLP modules, and the role of the warm-up phase. Detailed results and discussions are provided in Appendix C.

\subsection{Versatility Study}
We evaluate the effectiveness of MnLP on another SL-based policy (i.e., SIL~\cite{luo2025boosting}) in Appendix C, where we demonstrate its versatility on different neural architectures.

\section{Conclusion}
\label{sec:conc}

We propose MnLP, a training-only multi-depth supervision scheme for SL-based neural routing policies that adds a lookahead mechanism to predict future nodes, injecting long-horizon signals without changing inference or adding test-time cost. Experiments show consistent gains across cross-size, cross-distribution, and real-world benchmarks, highlighting the value of multi-step supervision for NCO.

Future work includes extending MnLP to RL for long-horizon credit assignment, applying it to other CO tasks (e.g., scheduling), and designing test-time retained multi-step predictors to improve efficiency and solution quality.

% \section*{Ethical Statement}

% There are no ethical issues.

% \section*{Acknowledgments}

% The preparation of these instructions and the \LaTeX{} and Bib\TeX{}
% files that implement them was supported by Schlumberger Palo Alto
% Research, AT\&T Bell Laboratories, and Morgan Kaufmann Publishers.
% Preparation of the Microsoft Word file was supported by IJCAI.  An
% early version of this document was created by Shirley Jowell and Peter
% F. Patel-Schneider.  It was subsequently modified by Jennifer
% Ballentine, Thomas Dean, Bernhard Nebel, Daniel Pagenstecher,
% Kurt Steinkraus, Toby Walsh, Carles Sierra, Marc Pujol-Gonzalez,
% Francisco Cruz-Mencia and Edith Elkind.

\bibliographystyle{named}
\bibliography{ijcai26}

\appendix

\clearpage

\section{Computational Details}

In this section, we define the computations carried out within a single attention block in the decoder of the LEHD model, which is used as the backbone in our MnLP framework. The attention block basically adheres to the standard Transformer architecture, comprising a multi-head self-attention mechanism, residual connections, layer normalization, and a position-wise feed-forward network. These components collectively enable the model to capture rich contextual relationships across nodes of the VRP instances.

\paragraph{Multi-Head Self-Attention ($\operatorname{MultiHeadAttn}$)}
The self-attention mechanism enables each node to attend to all others in the input sequence, allowing the model to capture dependencies across different parts of the graph.

Given an input sequence $X \in \mathbb{R}^{n \times d}$, where $n$ is the number of nodes and $d$ is the embedding dimension, we compute the query, key, and value matrices as follows:
\[
Q = XW^Q, \quad K = XW^K, \quad V = XW^V
\]
where $W^Q, W^K, W^V \in \mathbb{R}^{d \times d_h}$ are learnable weight matrices, and $d_h$ is the dimension of each attention head.

For each of the $h$ heads, the attention output is computed via scaled dot-product attention:
\[
\text{head}_i = \text{Softmax} \left( \frac{Q_i K_i^\top}{\sqrt{d_h}} \right) V_i
\]

The outputs of all heads are concatenated and projected:
\[
\operatorname{MultiHeadAttn}(X) = \operatorname{concat}(\text{head}_1, \dots, \text{head}_h) W^O
\]
where $W^O \in \mathbb{R}^{hd_h \times d}$ is a learnable projection matrix.

\paragraph{Residual Connection}
To facilitate training and avoid vanishing gradients, a residual connection is added between the input and the multi-head attention output:
\[
X' = X + \operatorname{MultiHeadAttn}(X)
\]

\paragraph{Feed-Forward Network ($\operatorname{FFN}$)}
The FFN applies a two-layer feed-forward transformation independently to each position (i.e., node representation):
\[
\operatorname{FFN}(x) = \text{ReLU}(xW_1 + b_1) W_2 + b_2
\]
where $W_1 \in \mathbb{R}^{d \times d_{ff}}$ and $W_2 \in \mathbb{R}^{d_{ff} \times d}$ are learnable matrices, and $b_1, b_2$ are the bias terms. In our experiments, we set $d_{ff} = 512$.

\paragraph{Second Residual Connection}
The output of the FFN is added to its input through a second residual connection:
\[
\text{Output} = X' + \operatorname{FFN}(X')
\]

This output serves as the representation passed to the next decoder layer or used in the downstream prediction head.

\vspace{3em}

\section{Cross-distribution Data Settings}

\begin{figure}[thb]
\centering
    % \subfloat[]{\includegraphics[width=0.22\textwidth]{imgs/tsp1000_rotation_n1000_fake_instance_1_visualization.pdf}}\hfill
    % \subfloat[]{\includegraphics[width=0.22\textwidth]{imgs/tsp1000_explosion_n1000_fake_instance_2_visualization.pdf}}
    \includegraphics[width=0.5\textwidth]{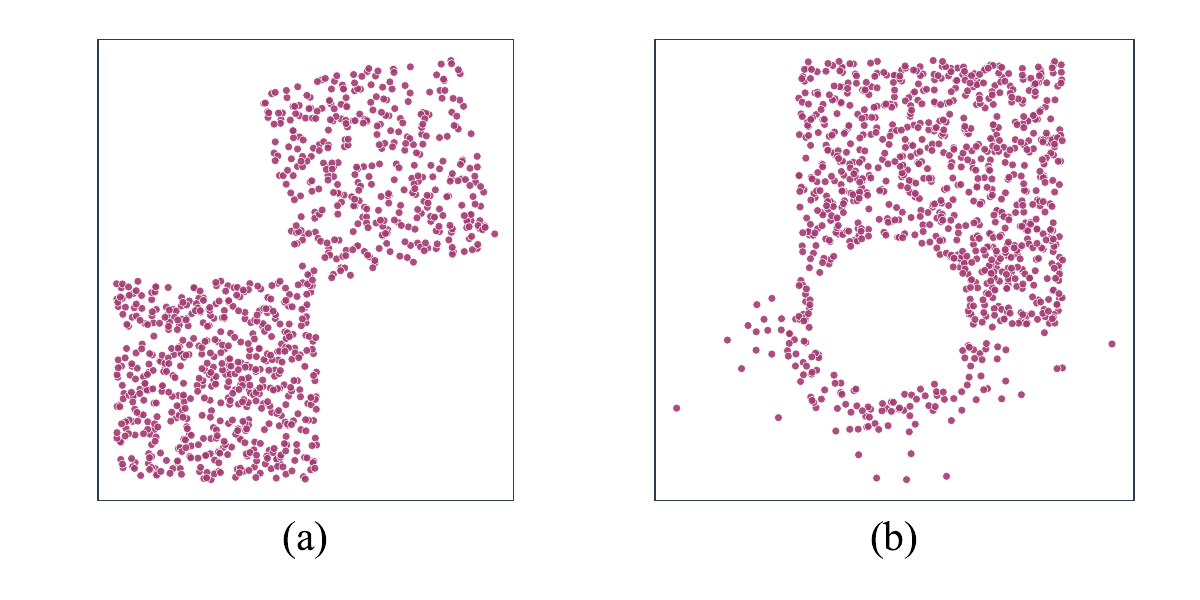}
\caption{TSP1000 instances with different distributions. (a) Rotation; (b) Explosion.}
\label{fig:dist}
\end{figure}

Following the previous studies \cite{bossek2019evolving,omni2023}, we mutate nodes that originally follow the uniform distribution to create the rotation and explosion data distributions. The example instances for these two distributions are visualized in Figure~\ref{fig:dist}.

Specifically, for rotation distribution, we mutate nodes by rotating a subset around the origin. Selected node coordinates are transformed using the rotation matrix $\begin{bmatrix} \cos(\varphi) & -\sin(\varphi) \\ \sin(\varphi) & \cos(\varphi) \end{bmatrix}$ with rotation angle $\varphi \sim [0, 2\pi]$.

For explosion distribution, we mutate uniformly distributed nodes by simulating a random explosion. We randomly select an explosion center $\mathbf{v}_c$ and move all nodes $\mathbf{v}_i$ within radius $R=0.3$ away from the center using $\mathbf{v}_i \leftarrow \mathbf{v}_c + (R+s)\frac{\mathbf{v}_i-\mathbf{v}_c}{\lVert\mathbf{v}_i-\mathbf{v}_c\rVert}$, where $s\sim \text{Exp}(\lambda=1/10)$.

\section{More Experiment Results}

\paragraph{Comparison Study with RRC.} Since the LEHD model can typically be enhanced during inference via Random Re-construction (RRC)~\cite{luo2023neural}, we also compare our MnLP-trained model with the original LEHD model under the same number of RRC steps (i.e., 1000). As shown in Table~\ref{tab:comparison}, the MnLP-trained models outperform the original LEHD models, particularly on large-scale problems (i.e., $n=500$ and $n=1000$). However, on smaller-scale problems such as CVRP100 and CVRP200, our model performs slightly worse. One possible explanation is that RRC is primarily designed to escape local optima and alleviate myopic behavior by reconstructing solutions, whereas our MnLP training already equips the model with long-horizon planning capabilities. As a result, the additional benefit from RRC on small-scale instances is diminished.

\begin{table*}[htbp]
\centering
{\fontsize{9}{11}\selectfont
\begin{tabular}{@{}llcccccccc@{}}
\toprule
\multirow{2}{*}{Problem} & \multirow{2}{*}{Method} & \multicolumn{2}{c}{$n=100$} & \multicolumn{2}{c}{$n=200$} & \multicolumn{2}{c}{$n=500$} & \multicolumn{2}{c}{$n=1000$} \\
\cmidrule(lr){3-4} \cmidrule(lr){5-6} \cmidrule(lr){7-8} \cmidrule(lr){9-10}
& & Obj. & Gap & Obj. & Gap & Obj. & Gap & Obj. & Gap \\
\midrule
\multirow{2}{*}{TSP} 
& LEHD-RRC 1000 & 7.763 & 0.002\% & 10.705 & 0.017\% & 16.554 & 0.197\% & 23.298 & 0.771\% \\
& Ours-RRC 1000 & 7.763 & 0.001\% & 10.705 & 0.017\% & 16.550 & 0.174\% & 23.268 & 0.639\% \\
\midrule
\multirow{2}{*}{CVRP} 
& LEHD-RRC 1000 & 15.630 & -0.105\% & 20.103 & -0.343\% & 37.138 & -0.245\% & 37.728 & 1.719\% \\
& Ours-RRC 1000 & 15.648 & 0.012\% & 20.121 & -0.255\% & 37.133 & -0.257\% & 37.516 & 1.148\% \\
\bottomrule
\end{tabular}
}
\caption{Performance comparison between models with RRC 1000 on TSP and CVRP.}
\label{tab:comparison}
\end{table*}

\begin{table}[t!]
\centering
\setlength{\tabcolsep}{1.6mm}
{\fontsize{9}{11}\selectfont
\begin{tabular}{lccccc}
\toprule
 & $K=0$ & $K=1$ & $K=2$ & $K=3$ & $K=4$  \\ 
\midrule
Time    & 23.5 & 26.1  & 28.2  & 30.9  & 33.0 \\
Memory    & 55.3 & 56.2  & 57.1 & 58.0 & 58.8 \\
\bottomrule
\end{tabular}
}
\caption{The training time costs (min.) and model sizes (MB) of MnLP models with different numbers of MnLP modules. The experiments are based on TSP100 with a batch size of 32.}
\label{tab:complexity}
\end{table}

\begin{figure}[thb]
\centering
    \subfloat[]{\includegraphics[width=0.24\textwidth]{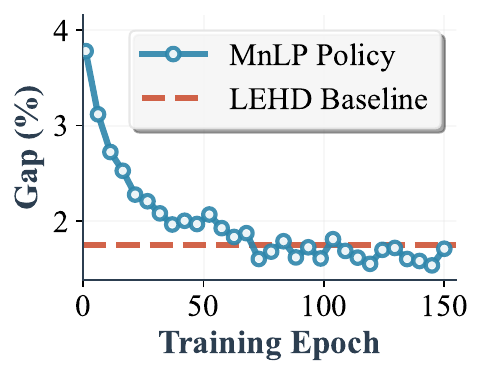}}\hfill
    \subfloat[]{\includegraphics[width=0.24\textwidth]{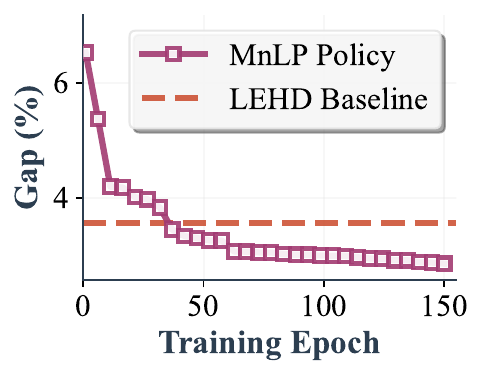}}
\caption{The model performance on TSP instances with different training epochs. (a) TSP500; (b) TSP1000.}
\label{fig:converge}
\end{figure}

\paragraph{Computational effect of MnLP training.} Training neural routing policies with the proposed strategy introduces additional computational overhead compared to training the main model alone. Thus, we empirically analyze the additional training overhead introduced by MnLP training with varying numbers of MnLP modules.  Taking TSP100 as a case study, we vary the number of MnLP modules and report both the training time per epoch and the GPU memory usage (i.e., the peak GPU memory during forward passes) in Table~\ref{tab:complexity}. It can be found that adding a single MnLP module results in only a marginal increase in time and memory cost compared to the original LEHD model (i.e., $K=0$).

\begin{figure}[th]
\centering
    \subfloat[]{\includegraphics[width=0.42\textwidth]{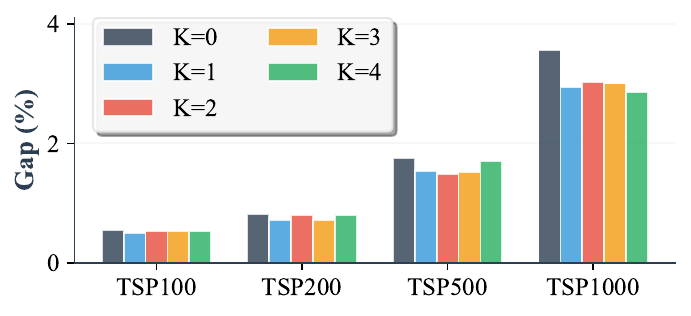}}\hfill
    \subfloat[]{\includegraphics[width=0.42\textwidth]{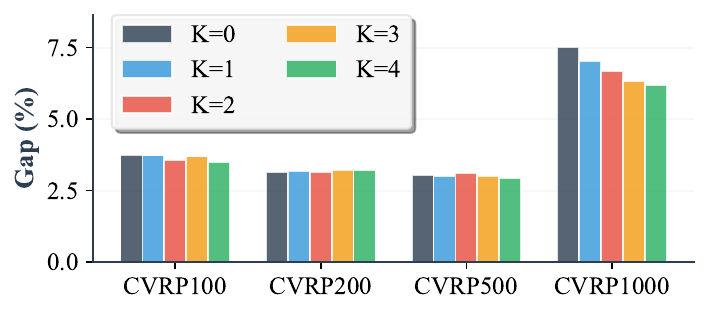}}
\caption{Performance comparison of the LEHD model with different numbers of MnLP modules (i.e., the value of $K$).}
\label{fig:diff_k}
\end{figure}

While training with multiple MnLP modules (e.g., $K=4$) introduces slightly higher computational cost, it facilitates faster and more effective convergence. To evaluate convergence behavior, we set $K=4$ and save the MnLP-trained models at every 5 training epochs. These models are then evaluated on TSP500 and TSP1000 instances, with results presented in Figure~\ref{fig:converge}. As shown in the figure, the MnLP-trained policy outperforms the original LEHD model on the TSP500 task after only 75 training epochs, and surpasses the baseline on TSP1000 in fewer than 50 epochs. This demonstrates that MnLP not only improves final performance but also accelerates the convergence of training. Despite the increased training time per epoch introduced by MnLP, it ultimately achieves better performance with reduced total training time. This conclusion also holds for the CVRP task, as demonstrated in Table 1, where our MnLP-trained model surpasses the LEHD baseline after only 15 training epochs, compared to 40 epochs required by the baseline.

\begin{table}[thbp]
\centering
\setlength{\tabcolsep}{1.6mm}
{\fontsize{9}{11}\selectfont
\begin{tabular}{c|l|c|c|c|c}
\toprule
 \multicolumn{2}{c|}{Size ($n$)}       & $100$ & $200$ & $500$ & $1000$  \\ 
\midrule
\multirow{2}{*}{TSP}
&w. WU    & 0.531\% & 0.795\% & 1.711\%  & 2.852\%  \\
&w/o. WU     & 0.562\%& 0.819\%  & 1.718\%  & 3.000\%  \\
\midrule
\multirow{2}{*}{CVRP}
&w. WU    & 3.509\%& 3.206\%  & 2.928\%  & 6.195\%  \\
&w/o. WU     & 3.726\%& 3.313\%  & 3.448\%  & 6.737\%  \\
\bottomrule
\end{tabular}
}
\caption{The ablation study of the warm-up (WU) phase.}
\label{tab:warmup}
\end{table}

\paragraph{Effect of different numbers of MnLP modules.} 
We vary the value of $K$ (i.e., the number of MnLP modules) and train the model accordingly. The results for both TSP and CVRP tasks are visualized in Figure~\ref{fig:diff_k}. It can be observed that incorporating different numbers of MnLP modules can improve model performance to various extents, with larger values of $K$ generally yielding better results on large-scale problems (e.g., $n=1000$). This suggests that deeper lookahead supervision is more beneficial when tackling large instances by providing longer-range context. Meanwhile, increasing the number of MnLP modules can lead to additional computational cost (see Table~\ref{tab:complexity}); therefore, it is also important to strike a balance between performance and computational complexity during implementation.

\begin{table}[thbp]

\centering
\setlength{\tabcolsep}{1.6mm}
{\fontsize{9}{11}\selectfont
\begin{tabular}{c|c|c|c}
\toprule
 \multirow{2}{*}{Method}   & $n=1000$ & $n=5000$ & $n=10000$   \\ 
 \cmidrule(lr){2-4}
   & Obj. (Gap) &  Obj. (Gap) &  Obj. (Gap)   \\ 
\midrule
 SIL   & 23.57 (1.95\%) & 52.59 (3.17\%)  & 74.69 (4.05\%)   \\
 SIL (MnLP)    & 23.56 (1.89\%)& 52.55(3.09\%)  & 74.27 (3.47\%)   \\
\bottomrule
\end{tabular}
}
\caption{The comparison between the original SIL and SIL trained through MnLP. The averaged objective values and corresponding optimality gaps are reported.}
\label{tab:sil}
\end{table}

\paragraph{Effect of the warm-up phase.} We introduce a warm-up phase during training, in which $\gamma$ is gradually increased. This strategy is designed to prevent auxiliary supervision from overwhelming the learning dynamics of the main model in the early stages. As shown in Table~\ref{tab:warmup}, models trained without this phase exhibit consistently degraded performance, highlighting the importance of progressively incorporating the MnLP loss to stabilize and enhance the overall training process.

\paragraph{Versatility study.} Beyond LEHD, we assess the portability of MnLP on a distinct SL-based policy, SIL~\cite{luo2025boosting}, which differs architecturally and procedurally. SIL employs a linear cross attention within a Transformer-like neural model and leverages a self-improvement training schedule for better solving large-scale VRP instances. MnLP integrates into SIL as a purely training-time addition: we attach multiple MnLP modules and their losses to the shared backbone, while keeping the SIL decoder and inference procedure unchanged. Consequently, test-time complexity and latency are identical to the original SIL method.

Following the protocol in~\cite{luo2025boosting}, we use $K{=}4$ with a loss weight $\gamma{=}0.2$, and otherwise adopt the same self-improvement schedule and optimization hyperparameters. We evaluate on large-scale TSP with $n\in\{1000, 5000, 10000\}$. As summarized in Table~\ref{tab:sil}, MnLP yields consistent improvements across all sizes. In terms of optimality gap, the reductions are modest at smaller scales: $1.95\%\!\rightarrow\!1.89\%$ for $n{=}1000$ and $3.17\%\!\rightarrow\!3.09\%$ for $n{=}5000$, and become more pronounced at $n{=}10000$, where the gap drops from $4.05\%$ to $3.47\%$.

These results support two takeaways. First, MnLP is architecture-agnostic: it provides gains when layered onto a model with linear cross-attention and the SIL paradigm, without modifying the decoding policy or adding inference overhead. Second, the benefits grow with problem size, aligning with the intuition that multi-depth supervision strengthens long-horizon planning signals in the shared representation. Together, this evidences MnLP’s versatility as a drop-in training method for SL-based neural routing policies.

\end{document}